\documentclass[times,twocolumn,final,authoryear]{elsarticle}

\usepackage{prletters}
\usepackage{framed,multirow}

\usepackage{amssymb}
\usepackage{latexsym}

\usepackage{url}
\usepackage{xcolor}
\definecolor{newcolor}{rgb}{.8,.349,.1}
\usepackage{graphicx}
\usepackage[cmex10]{amsmath}
\usepackage{mathtools}
\usepackage{enumerate}
\usepackage{multirow}
\usepackage{booktabs}
\usepackage{pifont}
\newcommand{\cmark}{\ding{51}}%


\usepackage[linesnumbered,ruled,vlined]{algorithm2e}
\usepackage{algpseudocode}

\begin{document}

\thispagestyle{empty}
               

\ifpreprint
  \setcounter{page}{1}
\else
  \setcounter{page}{1}
\fi

\begin{frontmatter}
\title{Hierarchical and Efficient Learning for Person Re-Identification}

\author{Jiangning \snm{Zhang}\corref{cor1}} 
\author{Liang \snm{Liu}\corref{cor2}} 
\author{Chao \snm{Xu}\corref{cor3}} 
\author{Yong \snm{Liu}\corref{cor4}} 


\received{1 May 2013}
\finalform{10 May 2013}
\accepted{13 May 2013}
\availableonline{15 May 2013}
\communicated{S. Sarkar}
\begin{abstract}
  Recent works in the person re-identification task mainly focus on the model accuracy while ignore factors related to the efficiency, e.g. model size and latency, which are critical for practical application. In this paper, we propose a novel \emph{\textbf{H}ierarchical and \textbf{E}fficient \textbf{Net}work} (HENet) that learns hierarchical global, partial, and recovery features ensemble under the supervision of multiple loss combinations. To further improve the robustness against the irregular occlusion, we propose a new dataset augmentation approach, dubbed Random Polygon Erasing (RPE), to random erase irregular area of the input image for imitating the body part missing. We also propose an \emph{\textbf{E}fficiency \textbf{S}core} (ES) metric to evaluate the model efficiency.  Extensive experiments on Market1501, DukeMTMC-ReID, and CUHK03 datasets shows the efficiency and superiority of our approach compared with epoch-making methods.
  \\
  \textit{Keywords}: Person Re-Identification; Hierarchical Features; Random Polygon Erasing; Efficient Score
\end{abstract}

\begin{keyword}
\MSC 41A05\sep 41A10\sep 65D05\sep 65D17
\KWD Person Re-Identification\sep Hierarchical Features\sep Random Polygon Erasing\sep Efficient Score
\end{keyword}
\end{frontmatter}

\section{Introduction}  \label{sec1}
\vspace{-6pt}
The person re-identification aims at retrieving corresponding images of a given person among the gallery person database, which has vast promising applications such as video surveillance and criminal investigation. Since \cite{first_DL} first apply the deep neural network to solve the ReID task, innumerous methods~\citep{zhang2017alignedreid,PCB,zhu2017part,HPM,MGN} emerge in succession. However, there are still challenges to apply the current person ReID methods into practical application. In the article, we focus on how to efficiently design and train the ReID model, and the idea can easily help to boost the performance of other methods.

Network architecture is one of the most concerned problems. \cite{pose_Zheng} and \cite{mask_Song} propose to use extra pose or mask information to improve the model performance, but they require an extra module to estimate the information and suffer from an extra time consumption.
\cite{attention_Liu} introduces an attention mechanism to enhance the model discrimination, and methods~\citep{GAN_Zhong,GAN_Liu} involve the idea of GAN to enrich the training dataset for reducing the impact of limited dataset. However, these methods generally require extra structures that increase the complexity of the network in practical application. Thus, many stripe-based methods are proposed, which are easy to follow and have pretty good performance. \cite{PCB} propose the PCB that divides the image into six stripes and obtains a good result. The popular MGN~\citep{MGN} acquires a better performance than other methods that uses multiple stripe branches.
However, nearly all current methods focus more on the accuracy while ignore the efficiency of the model, which is equally important for practical application. Considering the above reasons, we assimilate the stripe-designed idea and propose a novel \emph{\textbf{H}ierarchical and \textbf{E}fficient \textbf{Net}work} (HENet), which use multiple branches to efficiently learn hierarchical features.

\vspace{-1pt}
Some researchers focus on loss function designing, where stronger constraints can improve the model performance without increasing the model complexity. \cite{center} propose a center loss to learn the feature center of each class, while \cite{OIM} design a non-parametric OIM loss to leverage unlabeled data. \cite{triplet} apply the triplet loss to the ReID task and greatly improve the network performance by punishing intra-class and inter-class distances. Considering the design intention, we summarize center loss and OIM in one non-parametric category, while triplet and quadruplet losses in one hard sample mining category. Since different loss functions have different design intentions, we can use different kinds of loss functions simultaneously, expecting a mutual complementation among them for improving the model performance.

\begin{figure}[tb]
    \centering
    \includegraphics[width=0.49\textwidth]{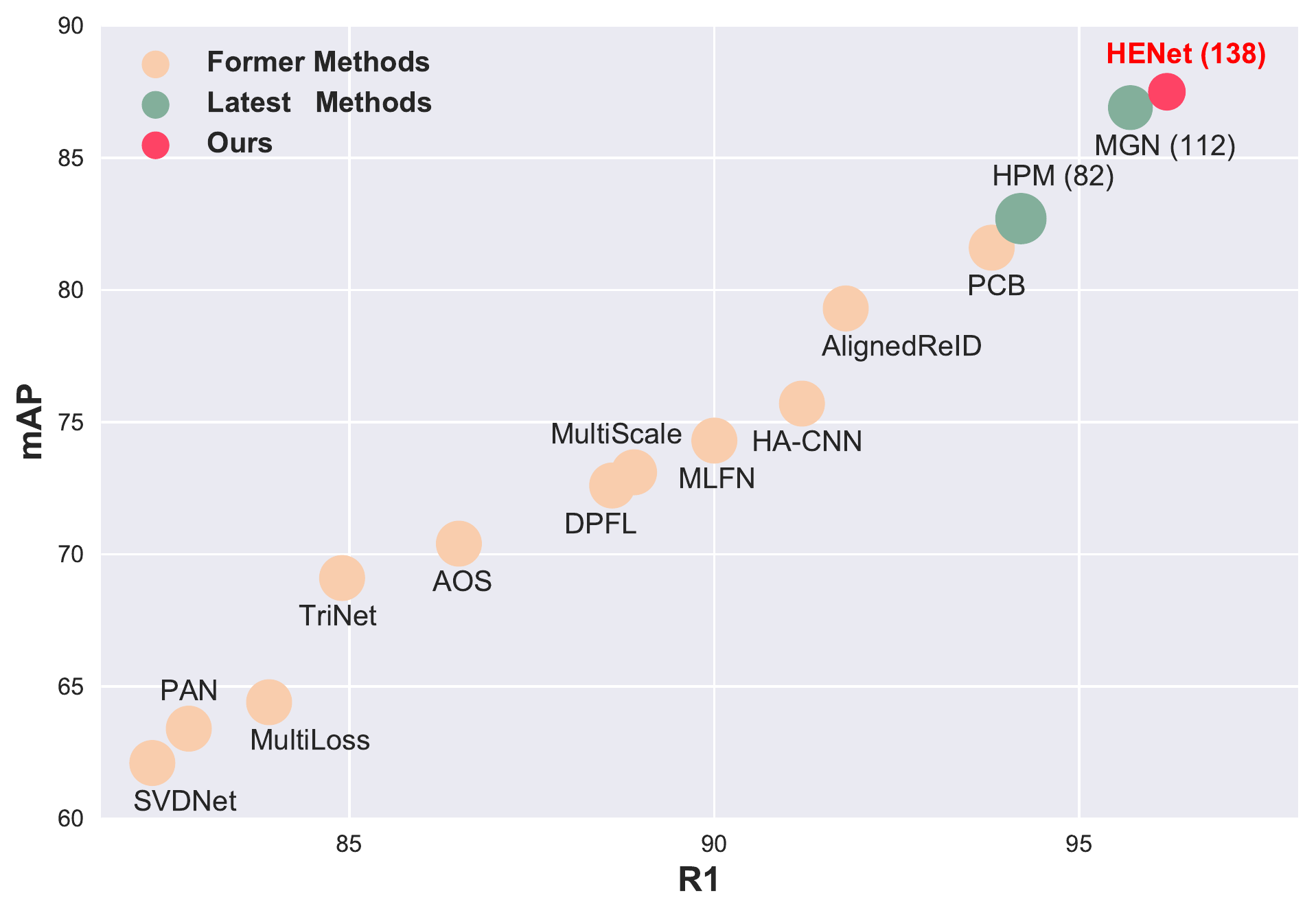}
    \caption{Performance of different epoch-making methods on Market1501. For our and latest popular methods, the marker size indicates the model size and the number in parenthesis indicates the running frame rate. Our proposed HENet outperforms all other methods in R1 and mAP, as well as has an equilibrium model size and a relatively faster running speed.}
    \label{fig:start}
    \vspace{-15pt}
\end{figure}

As we all know, the performance of deep neural network usually degrades in challenging scenarios, such as pose change, illumination intensity, and especially body occlusion. \cite{GAN_Liu} propose a pose-transfer network to extend the training dataset by generating images with different poses from only one input image, while \cite{GAN_Zhong} design a network named CamStyle to smooth the camera style disparities. \cite{RE} make a Random Erasing (RE) operation on the training dataset that effectively alleviate the occlusion. However, RE may not work well when confronted with some irregular object occlusions, such as backpacks and bicycles. So we propose a new \emph{\textbf{R}andom \textbf{P}olygon \textbf{E}rasing} (RPE) method to imitate the irregular occlusion, which improves the model performance.

To explore and solve above issues, we propose a novel HENet that learns global, partial, and newly designed recovery features simultaneously. During the training stage, we apply different loss function combinations to different branches, expecting a mutual complementation among different kinds of loss functions.
Furthermore, a novel RPE data augmentation method is proposed to boost performance, and we propose an \emph{\textbf{E}fficiency \textbf{S}core} (ES) metric to evaluate model efficiency. The designed HENet dose not need extra structure or operation during the testing stage, thus it is more suitable for practical application. Specifically, we make the following four contributions:

\begin{itemize}
\vspace{-6pt}
\item We propose an efficient HENet that learns hierarchical features ensemble under the supervision of multiple loss combinations, and it is more suitable for practical application than other methods.
\vspace{-6pt}
\item We propose a new data augmentation approach, dubbed RPE, to imitate the irregular occlusion, which greatly boosts the model performance.
\vspace{-6pt}
\item We propose an ES metric to evaluate the model efficiency for practical application.
\vspace{-6pt}
\item Our approach achieves a much better result than other epoch-making methods in three commonly used datasets, as well as acquires the highest efficiency score that shows its superiority for for practical industrial application, as shown in Fig.~\ref{fig:start}.
\end{itemize}

\section{Related Work}
\vspace{-6pt}
\subsection{Deep Person ReID}
Hand-crafted methods had been dominating the person ReID task until learning-based methods arrived. \cite{first_DL} first use neural network to solve the ReID issue which greatly improves the model performance.
Subsequent stripe-based methods~\citep{PCB,HPM,MGN} focus on learning local features, and they have a high accuracy and are efficient for employment. Works~\citep{pose_Zheng,mask_Song} leverage extra information, such as human pose and body mask, to improve the model performance, but they generally consume extra storage space and inference time. \cite{attention_Liu} introduce an attention mechanism to locate the active salient region that contains the person, while \cite{zhang2017alignedreid} and \cite{suh2018part} perform an alignment by calculating the shortest path between two sets of local features without requiring extra supervision. We employ stripe-based idea to design our model, which is easy to follow and has strong feature extraction ability for practical application.

\vspace{-6pt}
\subsection{Loss Function for Person ReID}
Cross-entropy loss is normally employed as a supervisory signal to train a classification model, and many researchers focus on designing the loss function for the ReID task. \cite{center} propose a center loss that simultaneously learns feature center of each class and penalizes different feature center, while \cite{OIM} propose an non-parametric OIM loss to further leverage unlabeled data. \cite{triplet} and \cite{quadruplet} employ a distance comparison idea idea to improve the network performance by punishing intra-class and inter-class distances. In the paper, we adopt multiple loss combinations in the training to extract hierarchical features ensemble, expecting a mutual complementation between different kinds of loss functions for improving the model performance.

\vspace{-6pt}
\subsection{Data Augmentation for Person ReID}
Some GAN-based methods normally contribute the model performance by enriching the training dataset. For instance, \cite{GAN_Zhong} use CycleGAN to style-transfer labeled images to other cameras, which reduces the over-fitting as well as increases the data diversity. 
\cite{GAN_Liu} propose the pose-transfer network to synthesize images with different poses from only one input image, while PTGAN~\citep{wei2018person} bridges the domain gap among datasets when transferring a person from one dataset to another. 
Besides the way of generating more diverse images, some related works aim at solving body occlusion by adding erasing operation. \cite{AE} use an adversarial erasing method to localize and expand object regions progressively. \cite{RE} propose a RE dataset augmentation approach, which first selects a random rectangle region in an image and then erases its pixels with fixed values. To reduce the impact of irregular body occlusions, we propose a novel \emph{random polygon erasing} (RPE) approach which brings a considerable improvement without adding complex structure or operation to the network.

\begin{figure*}[t]
  \centering
  \includegraphics[width=6in]{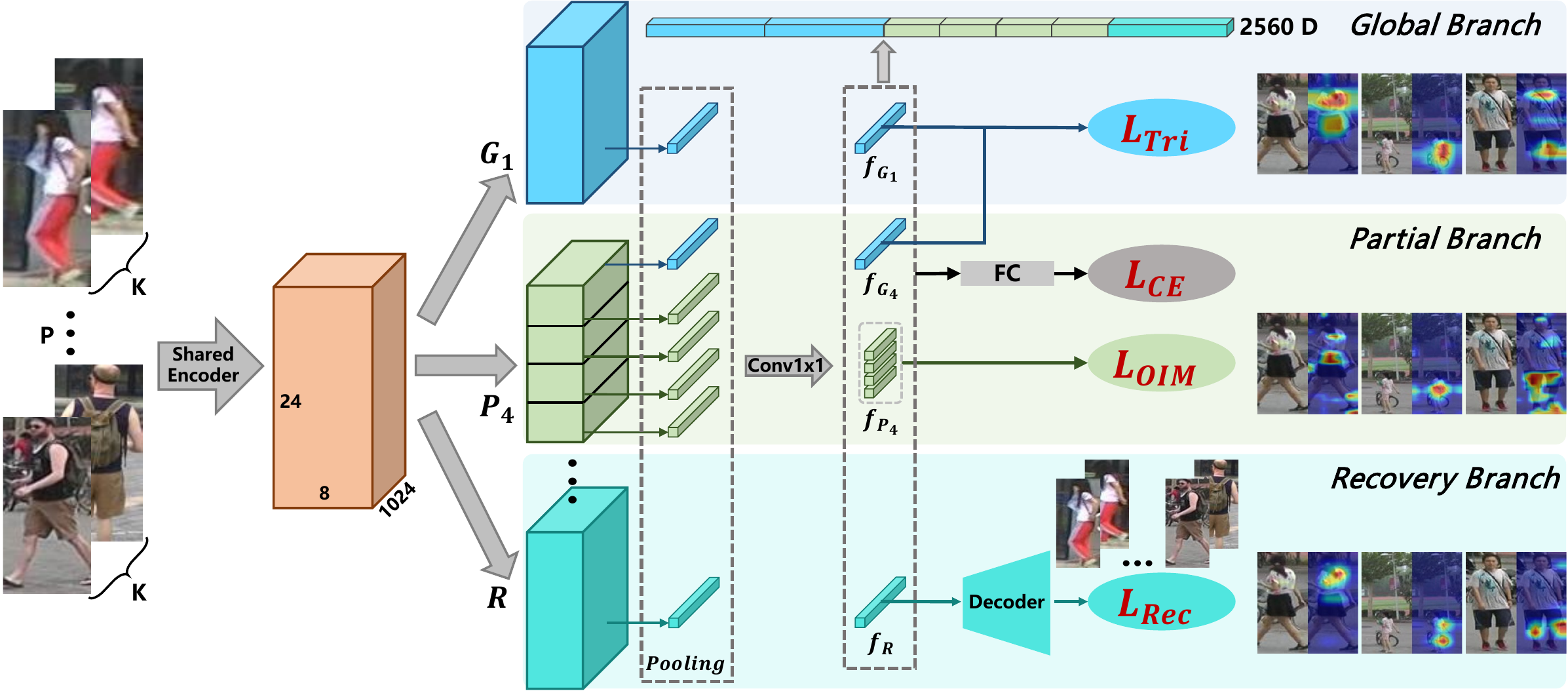}
  \caption{Diagram of the proposed HENet. The feature maps gone through a \emph{shared encoder} are split into $P_1$, $P_4$, and $R_{16}$ branches to further extract hierarchical features. During the training stage, global feature $f_G$ use triplet and CE losses as supervisory signals, partial feature $f_P$ use OIM and CE losses, and recovery feature $f_R$ uses Rec and CE losses. During testing stage, hierarchical features are concatenated to form the final representation. The right visual heatmaps are obtained by CAM~\citep{CAM} from three branches, where different branches focus on different regions.}
  \label{fig:HENet}
  \vspace{-10pt}
\end{figure*}

\section{Our Approach} \label{Our Approach}
\vspace{-3pt}
\subsection{The Structure of HENet} \label{p:HENet}
As shown in Fig.~\ref{fig:HENet}, the proposed HENet consists of three branches to extract hierarchical features. Specifically, global branch $G_1$ learns the global feature, while partial branch $P_4$ equally splits the whole feature map into 4 horizontal spatial bins for extracting partial features. Newly designed $R_{16}$ branch learns the recovery feature under the supervision of the reconstruction loss besides the CE loss.
Considering the efficiency and performance for practical application, we employ ResNet50~\citep{he2016deep} structure as the backbone of HENet that is line with other methods.

$GM_{1}$ is the global feature map belonging to branch $\boldsymbol{G_1}$ after convolution operations.
$GM_{4}$ and $PM_{4, j}$ are global and partial feature maps belonging to branch $\boldsymbol{P_4}$, where $PM_{4, j}$ is the $j_{th}$ bin that is split from $GM_{4}$.
Then we use pooling layer to generate global and partial intermediate features, $GI_{1}$, $GI_{4}$, and $PI_{4, j}$.

\vspace{-6pt}
\begin{equation}
GI_{i} = Pooling(GM_{i}), PI_{4, j} = Pooling(PM_{4, j}),
\end{equation}
where $i=1,4$ and $j=1,2,3,4$.
Subsequently, convolution with 1x1 kernel is employed to generate final global feature $\boldsymbol{f_{G_{i}}}$ with 512 vector dim and partial feature $\boldsymbol{f_{P_{4}}^{j}}$ with 256 vector dim.

\vspace{-6pt}
\begin{equation}
\boldsymbol{f_{G_{i}}} = Conv(GI_{i}), \boldsymbol{f_{P_{4}}^{j}} = Conv(PI_{4, j}).
\end{equation}
In the training stage, $\boldsymbol{f_{G_{i}}}$ are fed into linear layer $FC_{512}$ where Tri and CE losses serve as loss functions, while $\boldsymbol{f_{P_{4}}^{j}}$ are fed into linear layer $FC_{256}$ where OIM and CE losses are used.

$RM$ is the recovery feature map belonging to branch $\boldsymbol{R}$ after convolution operations, and we also use the pooling layer to generate recovery intermediate feature $RI$.

\vspace{-6pt}
\begin{equation}
RI = Pooling(RM).
\end{equation}
The convolution with 1x1 kernel is also employed to generate final recovery feature $\boldsymbol{f_{R}}$ with 512 vector dim.

\vspace{-8pt}
\begin{equation}
\boldsymbol{f_{R}} = Conv(RI).
\end{equation}
Then a decoder is applied to reconstruct a low-resolution image by inputting $\boldsymbol{f_{R}}$ during the training stage, where the reconstruction and CE losses serve as the cost loss functions.

\vspace{-6pt}
\subsection{Loss Functions} \label{Loss Functions}
During the training stage, we employ triplet and CE losses for the global branch, OIM and CE losses for the partial branch, as well as reconstruction and CE losses for the recovery branch.

\noindent\textbf{CE Loss}. The person ReID belongs to the multi-classification task, which is generally supervised by the CE loss function. Denote {$\boldsymbol{f}_{n}$} as the $n_{th}$ extracted feature that belongs to \{$\boldsymbol{f_{G_{1}}}$, $\boldsymbol{f_{G_{4}}}$, $\boldsymbol{f_{P_{4}}^{j}}$, $\boldsymbol{f_{R}}$\}, we obtain the following equation:

\vspace{-6pt}
\begin{equation} \label{eq:CE}
\mathcal{L}_{CE}=-\sum_{n=1}^{N}\log\frac{\exp(\boldsymbol{W}_{y_{n}}\boldsymbol{f}_{n})}{\sum_{j=1}^{C}\exp(\boldsymbol{W}_{j}\boldsymbol{f}_{n})},
\end{equation}
where $N$ is the batch size, $y_{n}$ is the real label of the $n_{th}$ extracted feature inside $C$ classes, and $\boldsymbol{W}_{j}$ is the weight vector for class $j$.

\noindent\textbf{Triplet Loss}.
As for global branch, we apply extra triplet loss to improve the model performance by punishing intra-class and inter-class distances in the training stage.
Specifically, we first random sample $P$ classes and then random sample $K$ images from each class. For each anchor sample $\boldsymbol{x}_{a}^{i}$ that belongs to the class $i$, we select a positive sample $\boldsymbol{x}_{p}^{i}$ and a negative sample $\boldsymbol{x}_{n}^j$ that belong to class $i$ and $j(i\neq j)$ respectively within a batch. The detailed formula is shown below:
\vspace{-6pt}
\begin{align}  \label{eq:Tri}
  \mathcal{L}_{Tri}(\theta; \boldsymbol{x}) = {\sum\limits_{i=1}^{P} \sum\limits_{a=1}^{K}}
      \Big[m & + \hspace*{-5pt} {\max\limits_{p=1 \dots K} \hspace*{-5pt} D\left(\boldsymbol{x}^i_a, \boldsymbol{x}^i_p\right)} - \hspace*{-5pt} {\min\limits_{\substack{j=1 \dots P \\ n=1 \dots K \\ j \neq i}} \hspace*{-5pt} D\left(\boldsymbol{x}^i_a, \boldsymbol{x}^j_n\right)} \Big]_+,
\end{align}
where $m$ controls the margin between intra-class and inter-class.

\noindent\textbf{OIM Loss}. As for partial branch, we choose OIM loss to further enhance the discrimination of the extracted feature.
When the category of the training dataset is large and each class only contains few instances, merely using CE loss may result in large variance of gradients in the classifier matrix, thus the model may not learn effectively~\citep{OIM}. Non-parametric OIM loss leverages extra unlabeled data in the training stage, which can make this deficiency up to some extent. As a result, we apply both OIM and CE losses to the partial branch, expecting mutual complementation between them. The formula with the input feature $\boldsymbol{x}$ is shown below:
\vspace{-6pt}
\begin{equation} \label{eq:OIM-pi}
p_i=\frac{\exp(\boldsymbol{v}_i\boldsymbol{x}/\tau)}{\sum_{j=1}^P\exp(\boldsymbol{v}_j\boldsymbol{x}/\tau)+\sum_{k=1}^Q\exp(\boldsymbol{u}_k\boldsymbol{x}/\tau)},
\end{equation}
\vspace{-6pt}
\begin{equation} \label{eq:OIM}
\mathcal{L}_{OIM}=\mathrm{E}_x\left[\log p_i\right],
\end{equation}
where $P$ is total class number, $Q$ denotes the size of circular queue that only contains unlabeled data, $\boldsymbol{v}_j$ and $\boldsymbol{u}_k$ denote weight vectors, and $\tau$ is a temperature parameter that a higher value indicates a softer probability distribution.

\noindent\textbf{Reconstruction Loss}.
The recovery branch is designed by an adversarial idea. In detail, the CE loss induces the branch $R$ to learn a part of the body feature that is enough for classification, while additional reconstruction loss compels the branch $R$ to learn from the whole image. As the adversarial training goes on, the network learns the feature only from the human body while ignoring the background.
Specifically, the recovery feature $\boldsymbol{f_{R}}$ goes through the \emph{Decoder} to reconstruct a low-resolution image, which is then used to calculate the distance against the real image by the \emph{Mean Square Error} (MSE) loss.

\vspace{-6pt}
\begin{equation} \label{eq:MSE}
\mathcal{L}_{Rec}=\sum(\boldsymbol{\hat{y}}_i - \boldsymbol{y}_i)^{2},
\end{equation}
where $\boldsymbol{\hat{y}}_i$ and $\boldsymbol{y}_i$ denote pixel values of the reconstructed and real images respectively.

\begin{figure}[t]
    \centering
    \includegraphics[width=3.4in]{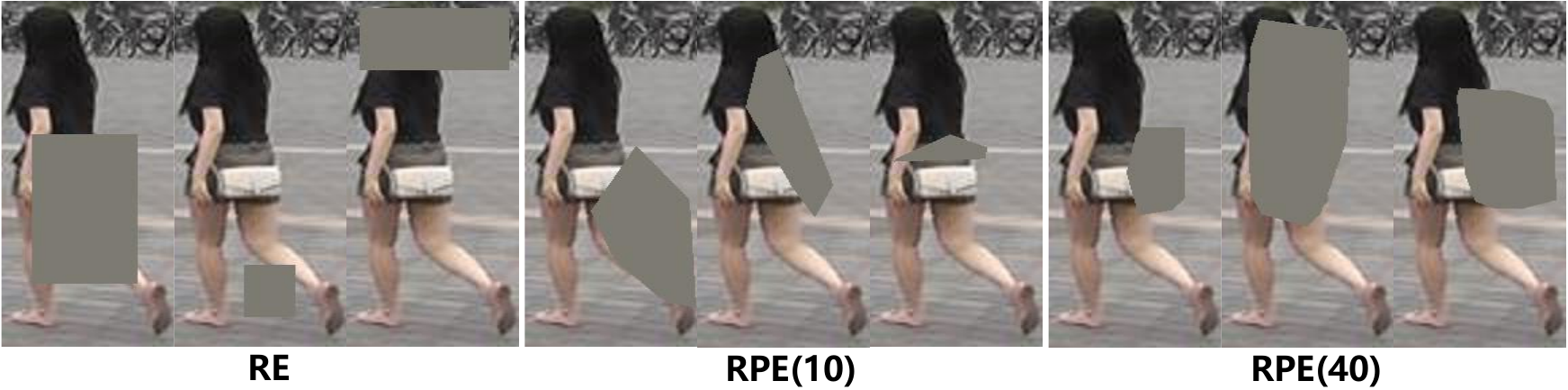}
    \caption{Visualization of RE and RPE operations. The gray indicates erased area and the number indicates the vertex number of the selected polygon.}
    \label{fig:RPE}
    \vspace{-15pt}
\end{figure}

\subsection{Random Polygon Erasing}\label{sec:RPE}
To reduce the impact of the irregular occlusion inside the person image, we propose a novel data augmentation method named RPE. As depicted in Fig.~\ref{fig:RPE}, the left column show the erasing results by RE~\citep{RE}, while the right two columns are from RPE with different vertex number, i.e. 10 and 40.
For an input image $\boldsymbol{I}$ with width $W$, height $H$, and area $S$, it has a probability $p$ to undergo the RPE operation. We first random generate the area ratio value $s_e$ between $s_l$ and $s_h$, as well as the aspect ratio $r_e$ between $r$ and $1/r$ ($r<1$). Then the erasing region with width $W_e=\sqrt{S\times s_e/r_e} (W_e<W)$ and height $H_e=\sqrt{S\times s_e\times r_e} (H_e<H)$ can be generated. Furthermore, a central reference point $P_c$ is randomly selected, whose coordinates $P_{c,x}$ and $P_{c,y}$ are in range $(W_e/2, W-W_e/2)$ and $(H_e/2, H-H_e/2)$ respectively. After that, $n$ points are randomly selected around the central reference point and the minimum external polygon mask can be calculated by generated points.
Finally, pixels of $\boldsymbol{I}$ that inside the mask are erased with a random value in range [0, 255].
The detailed procedure of RPE is showed in Algorithm \ref{algorithm_1}, and we set $p=0.5$, $s_l=0.02$, $s_h=0.45$, $r=0.35$ and $n=20$ in all experiments by default.

\begin{algorithm}[t]
\SetAlgoLined
\SetKwInOut{Input}{Input}
\SetKwInOut{Output}{Output}
\SetKwInput{Initialization}{Initialization}
\caption{Random Polygon Erasing Procedure}\label{algorithm_1}
\Input{
      Input image $\boldsymbol{I}$, Image size $W$ and $H$, Erasing probability $p$, Number of point $n$, Erasing area ratio range $s_l$ and $s_h$, Scale ratio $r$.}
\Output{Erased image $\boldsymbol{I^{\ast}}$.}
\Initialization{$p_1 \leftarrow $ Rand (0, 1), $cnt \leftarrow 1$.}

\eIf{$p_1 \geq p$}{
$\boldsymbol{I^{\ast}} \leftarrow \boldsymbol{I}$;\\
\Return{$\boldsymbol{I^{\ast}}$}.
}{
\While {True}{
  $S_e\leftarrow $ Rand $(s_l, s_h)\times S$; \\
  $r_e \leftarrow $ Rand $(r, 1/r)$; \\
  $W_e=\sqrt{S_e/r_e}$, $H_e=\sqrt{S_e\times r_e}$; \\
  \If{$W_e \le W$ and $H_e \le H$}{
    $P_{c,x} \leftarrow $ Rand $(W_e/2, W-W_e/2)$; \\
    $P_{c,y} \leftarrow $ Rand $(H_e/2, H-H_e/2)$; \\
    $w_{min} \leftarrow $ Max $(P_{c,x} - W_e/2, 0)$; \\
    $w_{max} \leftarrow $ Min $(P_{c,x} + W_e/2, W)$; \\
    $h_{min} \leftarrow $ Max $(P_{c,y} - H_e/2, 0)$; \\
    $h_{max} \leftarrow $ Min $(P_{c,y} + H_e/2, H)$; \\
    \While {cnt $\le$ p}{
      $p_{cnt,x}\leftarrow $ Rand $(w_{min}, w_{max})$;\\
      $p_{cnt,y}\leftarrow $ Rand $(h_{min}, h_{max})$;\\
      $cnt \leftarrow cnt + 1$;\\
      }
    $mask \leftarrow$ MinPolygon $([(p_{i, x}, p_{i, y})])$;\\
    $\boldsymbol{I}(mask) \leftarrow $ Rand $(0, 255)$;\\
    $\boldsymbol{I^{\ast}} \leftarrow \boldsymbol{I}$;\\
    \Return{$\boldsymbol{I^{\ast}}$}.
    }
  }
}
\end{algorithm}

\begin{table*}\setlength{\tabcolsep}{7pt}
  \centering
  \caption{Comparisons with epoch-making methods on three commonly used datasets. The right four columns are efficiency evaluations of different models.}
  \resizebox{.9\textwidth}{!}{
    \begin{tabular}{c|cc|cc|cc|cccc}
        \toprule
        \multirow{2}{*}{Methods} & \multicolumn{2}{c}{Market1501} & \multicolumn{2}{c}{DukeMTMC-ReID} & \multicolumn{2}{c}{CUHK03} & \multicolumn{4}{c}{Efficiency Evaluation}\\
        \cmidrule(l{3mm}r{3mm}){2-3} \cmidrule(l{3mm}r{3mm}){4-5} \cmidrule(l{3mm}r{3mm}){6-7} \cmidrule(l{3mm}r{3mm}){8-11}
        & R1 & mAP & R1 & mAP & R1 & mAP & FD $\downarrow$ & V(\emph{MB}) $\downarrow$ & S(\emph{FPS}) $\uparrow$ & ES $\uparrow$ \\
        \midrule
        SVDNet~\citep{sun2017svdnet} & 82.3 & 62.1 & 76.7 & 56.8 & 41.5 & 37.3 & - & - & - & -\\
        PAN~\citep{zheng2018pedestrian} & 82.8 & 63.4 & 71.6 & 51.5 & 36.3 & 34.0 & - & - & - & -\\
        MultiScale~\citep{chen2018person} & 88.9 & 73.1 & 79.2 & 60.6 & 40.7 & 37.0 & - & - & - & -\\
        HA-CNN~\citep{li2018harmonious} & 91.2 & 75.7 & 80.5 & 63.8 & 41.7 & 38.6 & - & - & - & -\\
        AlignedReID~\citep{zhang2017alignedreid} & 91.8 & 79.3 & 71.6 & 51.5 & 36.3 & 34.0 & 2048 & 100 & 207 & 2.55 \\
        PCB~\citep{PCB} & 93.1 & 81.0 & 82.9 & 68.5 & 63.7 & 57.5 & 1536 & 102 & 192 & \underline{3.45} \\
        HPM~\citep{HPM} & 94.2 & 82.7 & 86.6 & 74.3 & 63.1 & 57.5 & 3840 & 356 & 82 & 1.00 \\
        MGN~\citep{MGN} & 95.7 & 86.9 & 88.7 & 78.4 & 66.8 & 66.0 & 2816 & 263 & 112 & 2.42 \\
        \midrule
        HENet(Base) & 92.9 & 79.6 & 84.1 & 69.7 & 60.3 & 57.7 & 2048 & 169 & 160 & 2.76 \\
        HENet(+RPE) & 93.6 & 81.2 & 85.6 & 70.6 & 62.6 & 59.1 & 2048 & 169 & 160 & 3.08 \\
        HENet(+Losses) & \underline{95.6} & \underline{86.8} & \underline{88.9} & \underline{78.7} & \underline{66.5} & \underline{65.9} & 2560 & 224 & 138 & 3.41 \\
        HENet(+Both) & \textbf{96.0} & \textbf{87.2} & \textbf{88.9} & \textbf{78.9} & \textbf{67.1} & \textbf{66.5} & 2560 & 224 & 138 & \textbf{3.52} \\
        \bottomrule
    \end{tabular}
  }
  \label{tab:compare}
  \vspace{-10pt}
\end{table*}

\section{Experiment}\label{Experiment}
\subsection{Implementation Details}
We use pre-trained $ResNet\_50$ on ImageNet to initialize the HENet, and the \emph{Decoder} consists of 4 convolution-deconvolution groups.
During the training stage, we resize the input image to $384\times 128$, and form batches by first random sampling 4 classes and then random sampling 4 images for each class. Adam is used as the optimizer with parameter settings ($\beta_1=0.9$, $\beta_2=0.999$) and weight decay $5e^{-4}$. We train the model for 200 epochs, set the base learning rate to $2e^{-4}$, and decay the learning rate to a tenth when arriving 160 epochs and 200 epochs.
HENet is trained in a single TITAN X GPU on PyTorch framework~\citep{paszke2017automatic}.

\subsection{Dataset and Evaluation Protocol}
\textbf{Market1501} includes 32,668 images of 1,501 persons detected by the DPM from six camera views. This dataset is divided into the training set with 12,936 images of 751 persons, the testing set with 3,368 query images, and 19,732 gallery images of 750 persons.

\textbf{DukeMTMC-ReID} is a subset of the DukeMTMC dataset that contains 36,411 images of 1,812 persons from eight camera views. This dataset is divided into the training set with 16,522 images of 702 persons, the testing set with 2,228 query images, and 17,661 gallery images of 1110 persons.

\textbf{CUHK03} consists of 14,097 images of 1,467 persons from six camera views and has two annotation types: manually labeled bounding boxes and DPM-detected bounding boxes (used in the paper). This dataset is divided into the training set with 7,365 images of 767 persons, the testing set with 1,400 query images, and 5,332 gallery images of 700 persons.

\textbf{Protocols} we used to evaluate the model performance contain \emph{mean average precision} (mAP), as well as \emph{Cumulative Matching Characteristic} (CMC) at \emph{rank-1} (R1), \emph{rank-5} (R5), and \emph{rank-10} (R10), which can be used in both Market1501 and DukeMTMC-ReID. As for CUHK03, we further adopt the protocol proposed in~\cite{re-Ranking} where the experiments are conducted with 20 random splits for computing averaged performance. 
Moreover, we propose a new metric named \emph{Effciency Score} (ES) to evaluate the efficiency of the model for practical application, which considers another three factors besides R1 and mAP, i.e. feature dim (FD), model size (V), and forward speed (S).

\vspace{-6pt}
\subsection{Comparison with Epoch-Making Methods}
We compare the proposed HENet with some epoch-making methods that are published in recent two years on aforementioned datasets. As shown in Table~\ref{tab:compare}, HENet-base only contains global and partial branches, while other experiments of the proposed HENet are conducted by adding different components over HENet-base, e.g. RPE, Loss, and both components. All experiments of our method only use random horizontal flipping and random erasing data augmentation methods that are in accordance with other epoch-making methods.

From the comparison results, our proposed method obtains competitive results on three datasets. Specifically, the proposed HENet obtains 96.0\% R-1 for Market1501, 88.9\% R-1 for DukeMTMC-ReID, and 67.1\% R-1 for CUHK03, which exceeds all other methods. Besides, our approach outperforms others on mAP: 87.2\%, 78.9\%, and 66.5\% for Market1501, DukeMTMC-ReID, and CUHK03 respectively.
To further evaluate each aforementioned component, we conduct experiments on whether to add the component or not. As shown in Table~\ref{tab:compare} of the last four lines, either extra RPE data augmentation method or hierarchical losses improves the model performance, and the network can get the best performance when both the components are applied simultaneously.

We also propose a new metric, dubbed \emph{Efficiency Score} (ES), to evaluate the model efficiency. As shown in the last four columns of Table~\ref{tab:compare}, our method has the highest ES which means that HENet is more suitable for practical application, and the details of ES will be illustrated in Section~\ref{ES}.

\begin{table}[t]\setlength{\tabcolsep}{8pt}
    \centering
    \caption{Experimental results with different partial branch quantity.}
    \begin{tabular}{c|cccc|c}
        \toprule
        Structure & FD & R1 & R5 & R10 & mAP \\
        \midrule
        P(1) & 512 & 87.92 & 95.46 & 96.97 & 68.48 \\
        P(1,2) & 1536 & 90.15 & 95.52 & 97.08 & 70.03 \\
        P(1,2,3) & 2816 & 90.41 & 95.64 & 97.09 & 71.43 \\
        P(1,2,3,4) & 4352 & \textbf{90.81} & \textbf{95.84} & \textbf{97.48} & \textbf{72.39} \\
        \midrule
        P(1,3) & 1792 & 90.08 & 95.64 & 97.12 & 70.12 \\
        P(1,4) & 2048 & 90.17 & 95.68 & 97.30 & 70.39 \\
        P(1,6) & 2560 & 90.24 & 95.78 & 97.39 & 70.72 \\
        \bottomrule
    \end{tabular}
    \vspace{5pt}
    \label{tab:PB_quantity}
    \vspace{-10pt}
\end{table}

\vspace{-6pt}
\subsection{Ablation Study}
In this section, several ablation experiments are conducted to demonstrate the effectiveness of our approach. Note that all the following experiments are based on Market1501 dataset with the same settings and the re-ranking~\citep{re-Ranking} is not used for the fair comparison.

\noindent\textbf{Structure Analysis.}
Before doing the ablation study of HENet, we conduct a series of experiments with different partial branch quality to illustrate why we use $G_1$ and $P_4$ as the global and partial branches.
As shown in Table~\ref{tab:PB_quantity} of the top half, the network has a better performance with the increasing of the partial branch quantity, and nearly reaches the peak when quantity equals three. We choose partial branch number as two considering the performance and efficiency in practical application, where the network has a pretty good performance as well as a low \emph{Feature Dim} (FD).
On the bottom of the table, we show the experimental results with two partial branch quality under different configurations, and choose P(1,4) as our base model because of its satisfying performance and proper feature dim.

\begin{table}[t]\setlength{\tabcolsep}{8pt}
  \centering
  \caption{Ablation study of different components combination.}
  \resizebox{.46\textwidth}{!}{
    \begin{tabular}{ccc|cccc|c}
    \toprule
        $G_1$ & $P_4$ & $R_{16}$ & R1 & R5 & R10 & mAP & FD \\
        \midrule
        \cmark &        &        & 87.92 & 95.46 & 96.97 & 68.48 & 512 \\
              & \cmark &        & 89.67 & 95.81 & 96.97 & 69.55 & 1536 \\
              &        & \cmark & 84.29 & 93.65 & 95.96 & 58.69 & 512 \\
        \cmark & \cmark &        & 90.17 & 95.68 & 97.30 & 70.39 & 2048 \\
        \cmark &        & \cmark & 88.98 & 95.28 & 96.91 & 70.42 & 1024 \\
              & \cmark & \cmark & 90.15 & 95.89 & 97.16 & 70.34 & 2048 \\
        \cmark & \cmark & \cmark & \textbf{90.62} & \textbf{96.21} & \textbf{97.52} & \textbf{70.97} & 2560 \\
        \bottomrule
    \end{tabular}
  }
  \vspace{5pt}
  \label{tab:structure}
  \vspace{-10pt}
\end{table}

\begin{figure*}[htbp]
  \centering
  \includegraphics[width=7in]{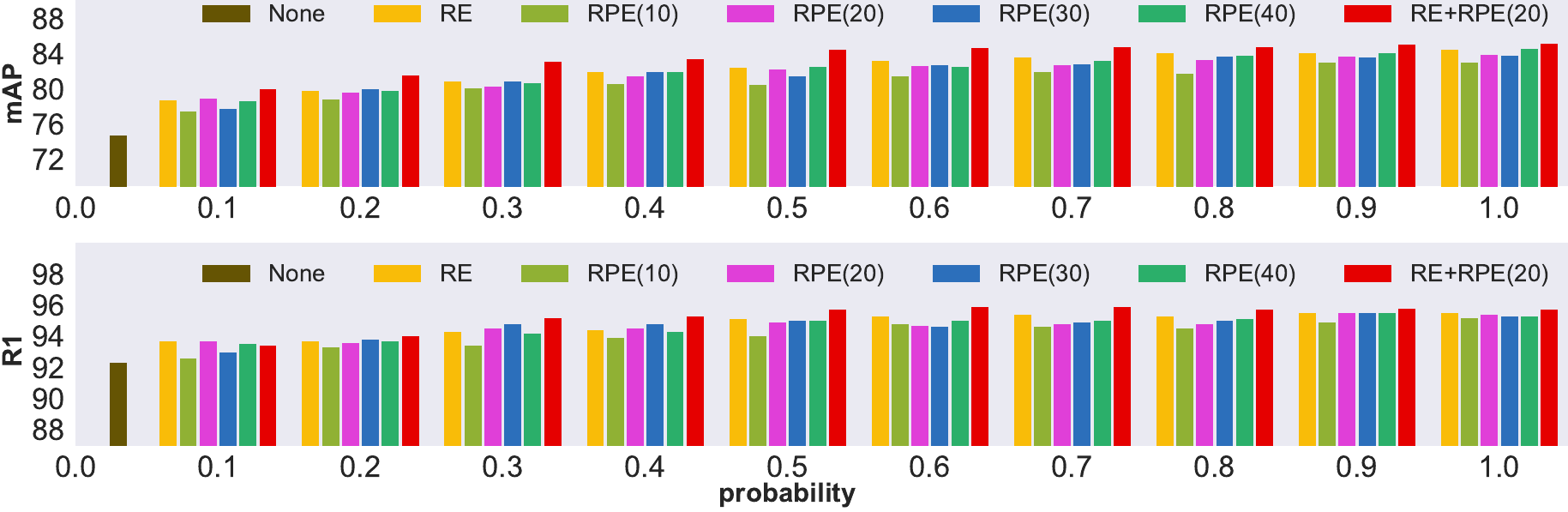}
  \vspace{-10pt}
  \caption{Metric evaluations (\%) of \emph{R1} and \emph{mAP} with different data augmentations on different levels of occlusion during the training stage in Market1501.}
  \label{fig:RPE1}
  \vspace{-2pt}
\end{figure*}

\begin{figure*}[htbp]
  \centering
  \includegraphics[width=7in]{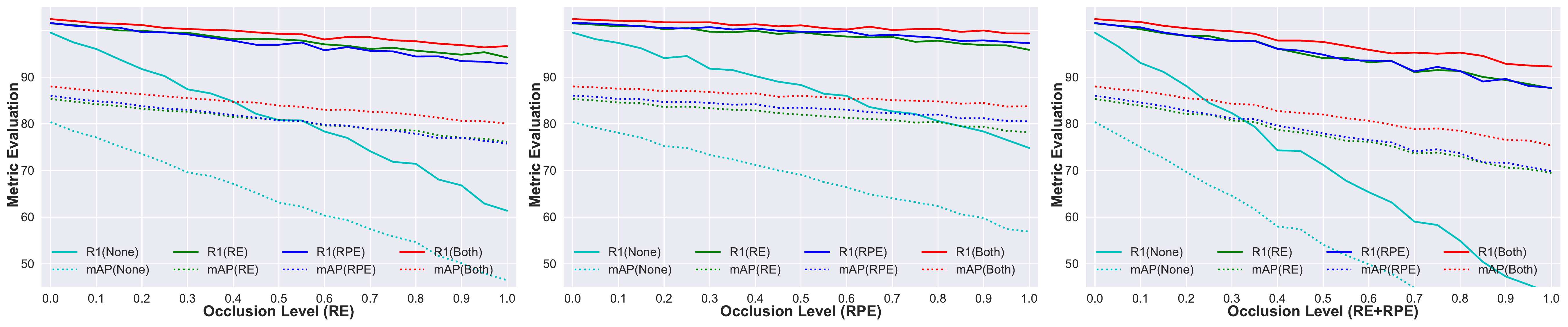}
  \vspace{-8pt}
  \caption{Metric evaluations (\%) on different levels of occlusion with different methods (From left to right are RE, RPE, and RE+RPE) in Market1501. The solid and dotted lines represent the evaluation results on \emph{R1} and \emph{mAP} with different data augmentation methods.}
  \label{fig:RPE2}
  \vspace{-10pt}
\end{figure*}

We further conduct an ablation to evaluate the effect of each branch. As shown in Table~\ref{tab:structure}, combining any two branches obtains a better performance than each single branch. 
When using all of three branches, the model achieves the highest score: R1=90.62\% and mAP=70.97\%, which demonstrates that each branch contributes the model performance.

\begin{table}[t]\setlength{\tabcolsep}{8pt}
  \centering
  \caption{Ablation study of different loss function combinations.}
  \resizebox{.46\textwidth}{!}{
    \begin{tabular}{cccc|cccc}
        \toprule
        & OIM & Tri & MSE & R1 & R5 & R10 & mAP \\
        \midrule
        \multirow{8}{*}{\rotatebox[origin=c]{90}{Market1501}}
        &        &        &        & 88.72 & 95.43 & 97.12 & 69.69 \\
        & \cmark &        &        & 91.33 & 96.02 & 97.74 & 73.42 \\
        &        & \cmark &        & 92.62 & 96.91 & 98.15 & 78.02 \\
        &        &        & \cmark & 90.43 & 95.90 & 97.18 & 71.65 \\
        & \cmark & \cmark &        & 92.90 & 97.30 & 98.25 & 80.04 \\
        & \cmark &        & \cmark & 91.65 & 96.56 & 97.83 & 74.37 \\
        &        & \cmark & \cmark & 93.20 & 97.68 & 98.40 & 79.90 \\
        & \cmark & \cmark & \cmark & \textbf{93.67} & \textbf{97.84} & \textbf{98.63} & \textbf{80.60} \\
        \bottomrule
    \end{tabular}
  }
  \vspace{5pt}
  \label{tab:Loss}
  \vspace{-10pt}
\end{table}

\noindent\textbf{Loss Settings.}
We conduct a group of experiments to evaluate each loss term. As shown in Table~\ref{tab:Loss}, using combined losses has a better performance than only single loss. Specifically, we can obtained the best result, i.e. R1=93.67\% and mAP=80.60\%, when all losses are applied, which illustrates that mixed loss functions are helpful for improving the model performance through mutual complementation.

\noindent\textbf{RPE Performance.}
To evaluate the effectiveness of RPE data augmentation method for the ReID task, we design a set of experiments under different probabilities with different experimental settings: RE, RPE(10), RPE(20), RPE(30), RPE(40), and RE+RPE(20), where the number is the vertex number (N) of the selected polygon. All experiments are based on the P(1,4) structure without using other data augmentation methods or tricks. As shown in Fig.~\ref{fig:RPE1}, the network performance gradually increases along with the probability, which can be consistently seen in both \emph{R1} (upper part) and \emph{mAP} (bottom part) metrics. Besides, the effectiveness of RPE increases when N is larger. We can observe that the network reaches a relatively good result when N=20 and a similar performance against RE when N=40 (The RPE degenerates into RE when N is large). Moreover, the experiment that combins two methods, marked as RE+RPE(20), is further conducted. The results indicate that this approach obtains a much better improvement than any single method, and it reaches a relatively peak state when \emph{probability equals 0.5} (R1=93.72\% and mAP=80.53\%).

To further show the robustness of \emph{Random Polygon Erasing} against occlusion, we add different occlusion levels and different data pre-processing methods, e.g. RE, RPE, and RE+RPE, to the test dataset in Market1501. As shown in Fig.~\ref{fig:RPE2}, the performance of all models decreases in both metrics with the increasing of the occlusion level, and the accuracy decreases faster when the test dataset is processed by RE+RPE (right sub-graph). 
The model trained with RPE (blue solid/dotted lines) or RE (green solid/dotted lines) has a stronger tolerance relative to the modified dataset than doing noting (cyan solid/dotted lines). And the model trained with both RPE and RE (red solid/dotted lines) outperforms others in a considerable margin, especially when the occlusion level is large. In short, results indicate that the RPE can greatly improve the robustness of the network, either alone or in conjunction with RE.

\subsection{Efficiency Evaluation}\label{ES}
To further evaluate the efficiency of the model for practical application, we calculate the \emph{Feature Dim} (FD), \emph{Model Volume} (V), and \emph{Forward Speed} (S) attributes for several models that have high scores in \emph{R1} and \emph{mAP}, as shown in Table~\ref{tab:compare}.
AlignedReID~\citep{zhang2017alignedreid} has the minimum model size and the maximum speed while HPM~\citep{HPM} is in contrast, and our model has an equilibrium size and a relatively faster speed. 
To evaluate model efficiency, we propose a new metric called \emph{Efficiency Score} (ES) which takes all aforementioned attributes into full account and can be used as the practical guideline.
Specifically, we first choose the reference models $M^1$ and $M^2$ that have the maximum size and minimum R1 score among comparison models. Then we calculate R1 and mAP scores for the comparison model $M^c$ in the following formula:

\vspace{-8pt}
\begin{equation}
\resizebox{.85\hsize}{!}{$
  Score_{M^c}^T=\frac{M^c_{V}\times {M^c_{S}}^2\times (M^c_{M}-M^2_{M}+thr)^3 / {M^c_{FD}}}{M^1_{V}\times {M^1_{S}}^2\times ({M^1_{M}-M^2_{M}+thr})^3 / {M^1_{FD}}}$,
}
\end{equation}
where $T$ denotes the category of metrics, $FD$ indicates feature dim, $V$ indicates model size, $S$ indicates forward speed, and $thr$ is the metric threshold that equals 30 in the paper. The formula fully considers multiple factors, e.g. \emph{R1}, \emph{mAP}, \emph{FD}, \emph{V}, and \emph{S}, which are important and must be considered on practical application. In general, $V\times S$ is close for different models, and extra $S$ term is multiplied to enhance the weight of the forward speed. Besides, considering the metric performance and the feature dim, we also add the latter two terms.
Final ES score of the comparison model $M^c$ can be obtained as follows:

\begin{equation}
ES_{M^c}=(Score_{M^c}^{R1}+\lambda Score_{M^c}^{mAP})/(1+\lambda),
\end{equation}
where $\lambda$ denotes weight and we set $\lambda$=1 in the paper.
As shown in the Table~\ref{tab:compare}, we calculate ES of AlignedReID, PCB, HPM, MGN, and our approach in Market1501 dataset, which are 2.55, 3.45, 1.00, 2.42, and 3.52 respectively. The results indicate that our approach has a higher \emph{ES} than other epoch-making methods and is superior for practical application.

\section{Conclusion}\label{Conclusion}
In this paper, we propose a novel end-to-end HENet for ReID task, which learns hierarchical global, partial, and recovery features ensemble. 
During the training stage, different loss combinations are applied to different branches for obtaining more discriminative features without increasing the model complexity. We further propose a new RPE data augmentation method to reduce the impact of irregular occlusions, which improves the performance and robustness of the network. Extensive experiments demonstrate the effectiveness and efficiency of our approach that is more suitable for practical application.

In the future, we will combine our approach with attention-based method to learn more discriminative features, as well as strengthen the recovery branch with the GAN idea to further improve the network performance.

\vspace{-8pt}

\bibliographystyle{model2-names}
\bibliography{prl_refs}


\end{document}